\documentclass[letterpaper, 10 pt, conference]{ieeeconf}  % Comment this line out if you need a4paper

\IEEEoverridecommandlockouts                              % This command is only needed if 
\usepackage{amsmath,amssymb,amsfonts}
\usepackage{algorithmic}
\usepackage{graphicx}
\usepackage{subfiles}
\usepackage{caption}
\usepackage{makecell}
\usepackage{tabularx}
\usepackage{multirow}

\usepackage{hyperref}
\usepackage{cleveref}
\crefformat{section}{\S#2#1#3} % see manual of cleveref, section 8.2.1
\crefformat{subsection}{\S#2#1#3}
\crefformat{subsubsection}{\S#2#1#3}

\newcolumntype{Y}{>{\centering\arraybackslash}X} % for table
\newcolumntype{b}{>{\hsize=1.2\hsize}X}
\newcolumntype{s}{>{\hsize=.5\hsize}X}
\newcolumntype{m}{>{\hsize=1.15\hsize}X}
\newcolumntype{B}{>{\hsize=0.8\hsize}X}
\newcolumntype{M}{>{\hsize=1.1\hsize}X}

\newcommand{\controlframe}{control coordinate system}
\newcommand{\controlframes}{control coordinate systems}

\newcommand{\Controlframes}{Control coordinate systems}
\newcommand{\ControlFrame}{Control Coordinate System}
\newcommand{\ControlFrames}{Control Coordinate Systems}

\title{\LARGE \bf
A Design Space of Control Coordinate Systems in Telemanipulation
}

\author{Yeping Wang$^{1}$, Pragathi Praveena$^{1}$ and Michael Gleicher$^{1}$% <-this % stops a space
\thanks{$^{1}$All authors are with the Department of Computer Sciences, University of Wisconsin-Madison, Madison, WI 53706, USA
        {\tt\small [yeping|pragathi|gleicher]@cs.wisc.edu}}%
\thanks{This work was supported by National Science Foundation award 1830242, Los Alamos National Laboratory and the Department of Energy.}% <-this % stops a space
}

\makeatletter
\let\@oldmaketitle\@maketitle% Store \@maketitle
\renewcommand{\@maketitle}{\@oldmaketitle% Update \@maketitle to insert...
   \vspace{5mm}
    \includegraphics[width=7in]{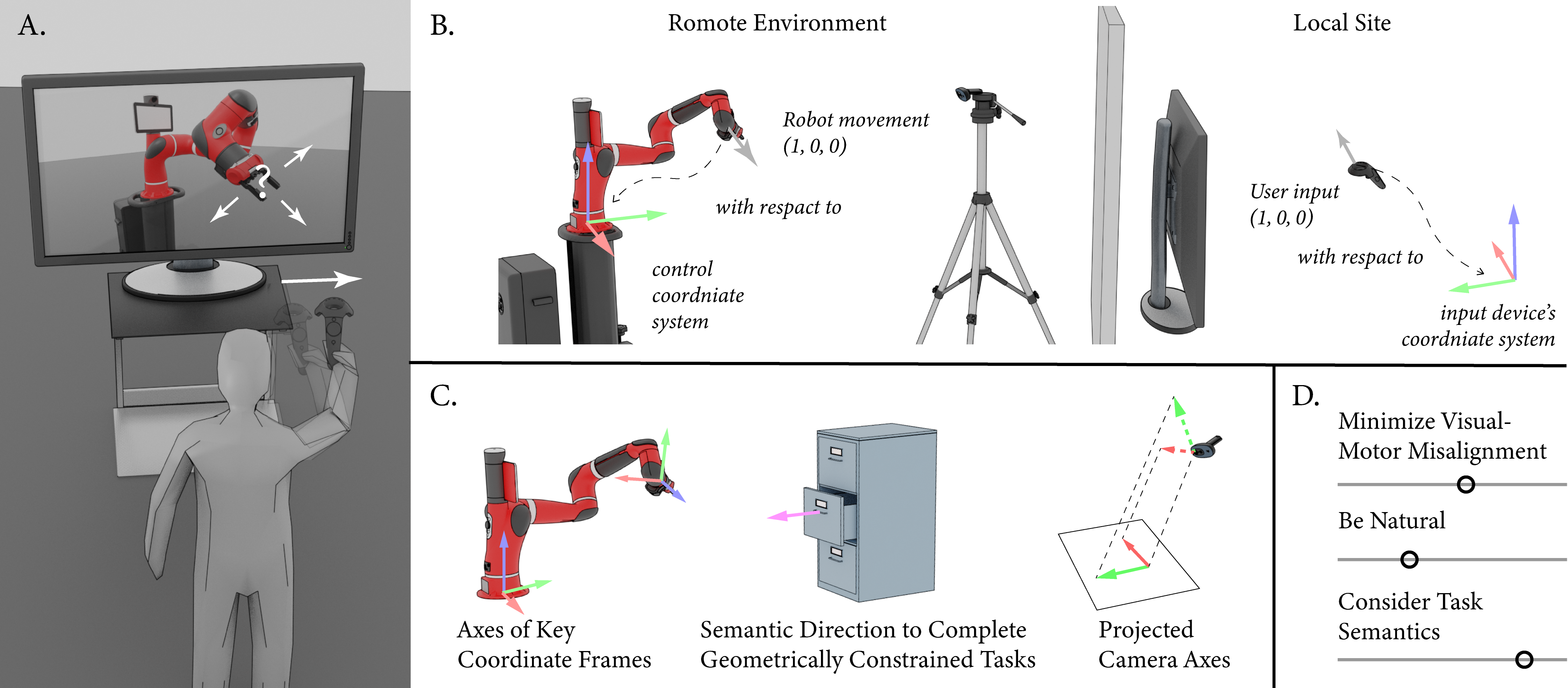}
    \captionof{figure}{\textbf{A.} Telemanipulation systems must determine how to map operator inputs to robot movements. \textbf{B.} \Controlframes{} can be selected in the remote environment to map user inputs to. In the figure, the \controlframe{} is a frame attached to the robot's base. A variety of other options may be suitable. \textbf{C.} In this paper, we enumerate common choices for the axes of \controlframes{}. \textbf{D.} Additionally, we present three design criteria for designers to understand the ramifications of the choices and reason about the suitability of \controlframes{} for different scenarios.}
    \label{fig: teaser}
    }
\makeatother

\begin{document}

\maketitle

\addtocounter{figure}{-1} % hack, otherwise following figures start with Fig 3

\thispagestyle{empty}
\pagestyle{empty}

%%%%%%%%%%%%%%%%%%%%%%%%%%%%%%%%%%%%%%%%%%%%%%%%%%%%%%%%%%%%%%%%%%%%%%%%%%%%%%%%
\begin{abstract}

Teleoperation systems map operator commands from an input device into some coordinate frame in the remote environment.
This frame, which we call a \textit{\controlframe{}}, should be carefully chosen as it determines how operators should move to get desired robot motions.
While specific choices made by individual systems have been described in prior work, a design space, \textit{i.e}., an abstraction that encapsulates the range of possible options, has not been codified.
In this paper, we articulate a design space of \controlframes{}, which can be defined by choosing a direction in the remote environment for each axis of the input device. Our key insight is that there is a small set of meaningful directions in the remote environment. \Controlframes{} in prior works can be organized by the alignments of their axes with these directions and new \controlframes{} can be designed by choosing from these directions. We also provide three design criteria to reason about the suitability of \controlframes{} for various scenarios. 
To demonstrate the utility of our design space, we use it to organize prior systems and design \controlframes{} for three scenarios that we assess through human-subject experiments.
Our results highlight the promise of our design space as a conceptual tool to assist system designers to design \controlframes{} that are effective and intuitive for operators.

\end{abstract}

%%%%%%%%%%%%%%%%%%%%%%%%%%%%%%%%%%%%%%%%%%%%%%%%%%%%%%%%%%%%%%%%%%%%%%%%%%%%%%%%

\section{Introduction}

% \begin{figure*} [bt]
%    \centering
%    \includegraphics[width=\textwidth]{figures/teaser.png}
%    %\vspace{-0.4cm}
%    \caption{\textbf{A.} Telemanipulation systems must determine how to map operator inputs to robot movements. \textbf{B.} \Controlframes{} can be selected in the remote environment to map user inputs to. In the figure, the \controlframe{} is a frame attached to the robot's base. A variety of other options may be suitable. \textbf{C.} In this paper, we enumerate common choices for the axes of \controlframes{}. \textbf{D.} Additionally, we present three design criteria for designers to understand the ramifications of the choices and reason about the suitability of \controlframes{} for different scenarios.}
    %\vspace{-0.4cm}
%    \label{fig: teaser}
%\end{figure*}

In telemanipulation, human operators and robots are generally in separated physical spaces. Operator commands are mapped from the input device into some coordinate frame in the robot's space. We call this frame the \textit{\controlframe} because operators perceive the robot as being controlled with respect to it. 
\Controlframes{} are a fundamental design choice for telemanipulation systems as they decide how operators should move to get desired robot motions. 
The design of \controlframes{} affects operators' sense of direction and spatial orientation in the remote environment.
However, there has been scant attention to the design of these \controlframes{}, with little systematic examination of the possibilities or comparison of their impact. Designers need to know what options they have and what criteria to follow to design \controlframes{} that are effective and intuitive for operators. 

%when choosing \controlframes{} for their telemanipulation systems.

In this paper, we provide a design space as a conceptual tool to articulate and explore the range of options for \controlframes{} (Figure \ref{fig: teaser}). By presenting a design space of the key choices in \controlframes{}, we provide a structure to organize and differentiate existing approaches and suggest new potential combinations. 
Our key insight is that there are a small number of meaningful directions in the remote environment. A \controlframe{} can be \emph{described} by how its axes align with these directions and \emph{designed} by choosing from these directions. 
Additionally, we provide three design criteria for designers to understand the ramifications of their choices and reason about the suitability of \controlframes{} for different scenarios. These criteria include considering whether a \controlframe{} causes visual-motor misalignments, is natural to human operators, and satisfies task semantics.

A design space can be evaluated according to its \emph{descriptive}, \emph{evaluative}, and \emph{generative} powers \cite{beaudouin2004designing}. To showcase the ability to describe existing \controlframes{} (descriptive power), we positioned existing \controlframes{} in our design space by describing how input device axes are mapped to the remote environment. To showcase the ability of our design space to assess design alternatives (evaluative power) and create
new designs (generative power), we conducted three case studies in which we apply the choices and reasoning afforded by the design space to design \controlframes{} and predict their suitability for three telemanipulation scenarios (\cref{sec:case_studies}). We assessed these designed \controlframes{} in human subject experiments to confirm our predictions (\cref{sec:evaluation}). Our results showed that the choices and reasoning afforded by our design space allowed us to design and predict suitable \controlframes{} for various scenarios.

The central contribution of this paper is a design space that consists of the key choices of \controlframes{} and design criteria to assist designers in reasoning about the suitability of \controlframes{} for different scenarios. By articulating a design space, we can design and assess various \controlframes{}, including \textit{hybrid frames} which leverage geometric information from the coordinate frames attached to two or more objects, providing alternative options for future teleoperation systems.

% In the remainder of this paper, we review the relevant prior research that motivates this work in \cref{sec:related_work}. We then describe our design space and use it to organize existing control coordinate systems in \cref{sec:design_space}. In \cref{sec:case_studies}, we provide three case studies in which we design and assess control coordinate systems with our design space. The designed control coordinate systems were evaluated in three independent user studies in \cref{sec:evaluation}. Finally, we conclude this paper with a discussion of the implications and limitations of this work in \cref{sec: discussion}.

\section{Related Work} \label{sec:related_work}
Our work builds upon prior work from three areas:  \controlframes{} in telemanipulation, visual-motor misalignments, and low-dimensional input devices for telemanipulation. 

\subsection{\ControlFrames{} in Telemanipulation}
While there are a variety of input devices for robot telemanipulation, in this paper we focus on spatial input devices that capture finger, hand or arm movements, such as computer mice, joysticks, or VR controllers. We do not consider non-spatial input devices such as sip-and-puff \cite{jain2019probabilistic} or electromyographic sensors \cite{artemiadis2007emg} in this paper because they offer a different mapping challenge as they must map non-spatial inputs to spatial outputs.

To control a robot manipulator's end-effector with a spatial input device, telemanipulation systems need to map user inputs to some \controlframe{} in the remote environment. 
The \controlframes{} in prior works fall into a few patterns, including a coordinate frame attached to the base of the robot \cite{hiatt2006coordinate, talha2019preliminary, rakita2018autonomous, praveena2022understanding}, the camera \cite{hiatt2006coordinate, wu2021camera, talha2019preliminary}, or the end-effector  \cite{hiatt2006coordinate, nawab2007joystick, chintamani2009improved, talha2019preliminary, wu2020development}. 
\textit{While specific choices made by individual systems have been described in prior works, this paper provides a design space for designers to understand the range of possible control coordinate systems and reason about their suitability.}

% learning-based mappings
Recently, learning-based methods have been employed to generate personalized mappings to compensate for individual differences \cite{li2020learning, pierce2012data} and task-specific mappings to leverage latent actions \cite{losey2020controlling, mehta2022learning}. 
These learning-based mappings follow manifolds derived from observations and are generally dynamic and non-linear. While they are not ``designed'' by choosing a \controlframe{}, these learning-based mappings can be interpreted and described using a dynamic \controlframe{} at each instant. Therefore, the design criteria afforded by our design space is still useful to understand and reason about learning-based mappings. 

\subsection{Visual-Motor Misalignments}
In telemanipulation systems, camera positions are selected to provide a clear view of the manipulation space. The camera viewpoint and robot \controlframe{} are not always aligned, which causes a misalignment between the robot motion on screen (visual) and the user's input motion (motor). This visual-motor misalignment increases the mental workload of human operators - they would need to use mental rotation strategies (spatial transformation of desired robot motions) or perspective-taking strategies (imagine how a scene looks like from another viewpoint) to move the robot \cite{menchaca2007influence}. 

To address visual-motor misalignments in robot telemanipulation, various solutions have been proposed including haptic devices \cite{kimmer2015effects}, visual cues \cite{nawab2007joystick, chintamani2009improved}, and training to improve the operator's spatial abilities \cite{menchaca2007influence}. 
However, these methods only \emph{help} operators handle visual-motor misalignments; to inherently eliminate visual-motor misalignments, the \controlframe{} should align with the coordinate frame of the camera. For example, DeJong \textit{et al.} \cite{dejong2011mental} present a mathematical framework that considers camera, display, and controller positions to minimize mental transformations in telemanipulation. 

While visual-motor misalignments affect teleoperation performance and human mental workload, they should not be the only factor to consider when choosing \controlframes{}. Ellis \textit{et al.} \cite{ellis2012human} find that the visual-motor misalignment disturbance is not linearly proportional to the degree of misalignment; human operators can adapt to some visual-motor misalignments. Such findings give designers more creative freedom, allowing them to choose \controlframes{} by considering other factors.\textit{In addition to minimizing visual-motor misalignments, our work identifies two other design criteria to assist designers in reasoning about and predicting the suitability of \controlframes{} (Section \ref{sec:criteria})}.
 
\subsection{Telemanipulation with Low-Dimensional Input Devices}
While many 3D input devices have been developed, low-dimensional input devices such as joysticks and computer mice are still widely used in professional teleoperation systems such as surgical robots \cite{gillen2014solo}, assistive robots \cite{herlant2016assistive}, disaster rescue robots \cite{xiao2021autonomous}, and space exploration robots\cite{pryor2020interactive}. 
Different kinds of 2D input devices have been used for a human operator to specify the position of a robot end-effector, including joysticks \cite{atherton2009supporting, baaberg2016adaptive, mower2019comparing, notheis2014evaluation}, computer mice\cite{leeper2012strategies}, and touch screens \cite{chung2017performance, hashimoto2011touchme}. 
Prior research has found low-dimensional input devices to be more accurate, comfortable, and easier to use for inexperienced operators over 3D input devices \cite{notheis2014evaluation, hachet2003cat, berard2009did, pryor2020interactive}. While having many benefits, low-dimensional input devices are intrinsically limited because they can only move an end-effector in a subspace of the 3D environment. Therefore, mode switching mechanisms \cite{herlant2016assistive, maheu2011evaluation} where operators switch the degrees of freedom of the robot that they want to control, and shared control methods \cite{abi2016visual} where operators control a subset of the robot's degrees of freedom and the remaining degrees of freedom are autonomously controlled by an autonomy system, are often employed for low-dimensional input devices. 
However, systems with mode switching or shared control methods still require mappings for user inputs where our design space can be applied.

Given the limitations of low-dimensional devices, their \controlframes{} should be carefully selected as the \controlframes{} represent which subspace the robot can move in. 
While many existing systems attach the \controlframe{} to the ground plane \cite{baaberg2016adaptive, talha2019preliminary, wu2020development} or the camera image plane \cite{atherton2009supporting, talha2019preliminary, wu2021camera}, several alternative choices may be appropriate.
\textit{In this paper, by articulating a design space of \controlframes{}, we introduce a hybrid frame that is specifically designed for low-dimensional input devices}.

\section{Design Space}  \label{sec:design_space}
In this section, we present our design space of control coordinate systems. 
Our design space \emph{describes} a \controlframe{} by how its axes align with meaningful directions identified by our design space. Our design space can be used to \emph{generate} a \controlframe{} by choosing a direction in the remote environment for each axis of the input device. Our design space also provides design criteria to \textit{evaluate} a \controlframe{}. The design criteria combine insights from prior research with our experience designing \controlframes{}.

This section follows the Questions-Options-Criteria framework \cite{maclean1991questions} to present our design space. First, we identify the key question in designing a \controlframe{} in Section \ref{sec:mappings}: designers need to determine how to map each axis of the input device to the remote environment. Then, in Section \ref{sec:choices}, we provide options that are possible answers to the question by enumerating common choices of axes. We use these options to organize \controlframes{} in the literature in Section \ref{sec:review}. Finally, Section \ref{sec:criteria} provides three design criteria to reason about and predict the suitability of \controlframes{} for a scenario.

\subsection{\ControlFrames}  \label{sec:mappings}

In telemanipulation systems, a spatial input device captures an operator's finger or arm movements and represents them in the input device's coordinate system. To map user inputs to robot motions, a coordinate system needs to be selected in the remote environment. The \controlframe{} can be described by the directions of its axes. Mathematically, we represent a \controlframe{} using a matrix where columns are unit vectors $\mathbf{d}_x, \mathbf{d}_y, \mathbf{d}_z \in \mathbb{R}^3$ that denote the directions of $x, y, z$ axes of the \controlframe{}:
\begin{equation} \label{eq: mapping_matrix} 
    \mathbf{T} = 
    \begin{cases}
    [\mathbf{d}_x, \mathbf{d}_y] & \text{2D input device}\\
    [\mathbf{d}_x, \mathbf{d}_y, \mathbf{d}_z] & \text{3D input device}
    \end{cases}
\end{equation}
The robot end-effector's linear velocity $\mathbf{v}_o \in \mathbb{R}^3$ and angular velocity $\boldsymbol{\omega}_o \in \mathbb{R}^3$ can be obtained by mapping translational and rotational user inputs $\mathbf{v}_i$ and $\boldsymbol{\omega}_i$ to the control coordinate system. 
The inputs $\mathbf{v}_i, \boldsymbol{\omega}_i \in \mathbb{R}^2$ on 2D input devices and $\mathbf{v}_i, \boldsymbol{\omega}_i \in \mathbb{R}^3$ on 3D input devices. 
\begin{equation} \label{eq: linear_transform} 
    [\mathbf{v}_o, \boldsymbol{\omega}_o] =  
    \mathbf{T} [\mathbf{v}_i, \boldsymbol{\omega}_i]
\end{equation}
User inputs $\mathbf{v}_i$ and $\boldsymbol{\omega}_i$ may have different physical meanings depending on the control method. \emph{Position control} is commonly used on VR controllers or computer mice, where a relative user movement is directly mapped to the robot. In position control, the $\mathbf{v}_i$ and $\boldsymbol{\omega}_i$ denote the operator's linear and angular velocity. Meanwhile, \emph{rate control} is commonly used on joysticks or space mice, where the displacement of the input device to the device's origin pose is mapped to the robot's velocity. In rate control, $\mathbf{v}_i$ and $\boldsymbol{\omega}_i$ denote the positional and rotational displacements. Specifically, $\boldsymbol{\omega}_i$ is a scaled axis whose direction is along the axis of a rotational displacement and whose norm is the rotation angle.

It is worth mentioning that a \controlframe{} does not have to be attached to a physical object. For example, the \controlframe{} 
 of \emph{orbit control} moves on an imaginary sphere (Figure \ref{fig: existing_frames}A). Moreover, because user inputs are mapped to the robot end-effector's velocity, the origin of a \controlframe{} does not affect the results of the mapping.

A \controlframe{} can be \emph{generated} by choosing a direction in the remote environment for each axis of the input device. We find that there is a relatively small set of common choices of these directions and list the common choices in Section \ref{sec:review}. To form a \controlframe{}, the direction for each input device axis may be chosen from the coordinate frame of the same object (\textit{e.g.}, the coordinate frame attached to the robot's base), or a designer may mix and match different directions. We call a \controlframe{} that leverages geometric information of two or more coordinate systems a \emph{hybrid frame}. In some cases, hybrid frames are created by choosing two directions and computing the third direction using the cross product. For example, a hybrid frame may be designed as the direction that moves the end-effector right in the camera image, the direction that moves the end-effector vertically up in the world frame, and the cross product of these two directions \cite{rakita2018autonomous, praveena2022understanding} (Figure \ref{fig: new_mappings}A). 
To maintain the generalizability of the design space, mutually orthogonal axes are not necessary for a \controlframe{}. For example, the \controlframe{} for a pouring task \cite{quere2020shared} in Figure \ref{fig: existing_frames}B maps 3D user inputs to two directions. One direction is vertical and moves the bottle up and down; the other direction is perpendicular to the bottle's axis and tilts the bottle. Although the two directions are not mutually orthogonal and don't span the 3D space, the \controlframe{} is sufficient to finish the pouring task.

\subsection{Choices for Directions } \label{sec:choices}

\Controlframes{} can be defined by choosing a direction in the remote environment for each axis of the input device. We organize the meaningful choices for directions into three categories. 

\emph{ The Axes of Key Coordinate Frames} --- 
The primary sources of axes used to map user inputs are the coordinate frames connected to the main entities: 
the global world frame, the coordinate frame attached to the robot's base, the frame of the robot's end-effector, the coordinate system defined by the camera, and the coordinate frame of the objects to be manipulated (\textit{i.e.}, the \emph{task frame} \cite{ballard1984task}, discussed as semantic directions below). 

The meaningfulness of each axis of these frames to operators may vary based on the scenario. 
For example, the \emph{upright} direction in the world frame aligns with gravity, which is essential to tasks like pouring or throwing.
Another example is that a camera's axes, which are determined by the camera orientation in the remote space, impact how the robot appears to move in the image captured by the camera. 
Moreover, the end-effector axes may be meaningful in describing motions relative to object interactions, such as approaching a grasp.

\emph{ Semantic Directions to Complete Geometrically Constrained Tasks } --- Many daily manipulation tasks are geometrically constrained, \textit{e.g.}, pulling a drawer, rotating a knob, or drawing on a table. 
A manipulation task's geometric constraints impose along which directions the robot may move to complete the task, \textit{e.g.}, the directions parallel to the table plane in a drawing task.
We define semantic directions as the directions to complete a geometrically constrained task. Semantic directions also include directions to complete softly constrained tasks such as the vertical direction to pick an object out of a box. 
Semantic directions can be specified manually or inferred from human demonstrations \cite{subramani2018inferring}.
When designing \controlframes{} for low-dimensional input devices, designers should consider mapping axes of the input device to these semantic directions. 
Low-dimensional input devices only move the robot in a subspace (\textit{e.g.}, a plane); selecting semantic directions ensures that operators can easily generate commands along the semantic directions. 

\emph{Projected Camera Axes} --- As described in Section \ref{sec:related_work}, selecting a camera frame as the \controlframe{} minimizes visual-motor misalignments. However, the camera frame is not a universally good design choice because it may be unnatural or lacking in semantics. 
For example, it is unnatural when horizontal human inputs are mapped to non-horizontal robot movement which can happen when a camera looks down at the robot and the optical axis is non-horizontal. In this case, if an operator pushes the input device horizontally away from them, the robot moves away from the camera along the non-horizontal optical axis of the camera.  
A \controlframe{} may lack task semantics when 2D user inputs are mapped to the camera image plane. This \controlframe{} only allows operators to freely move parallel to the image plane, which may not include the semantic directions related to a task. For example, in a tabletop drawing task, semantic directions are within the tabletop plane, but controlling a robot in the camera frame leads to movement parallel to the camera image plane.

An alternative design option is to project the camera axes onto a meaningful plane (\textit{e.g.}, the ground or a semantic plane to complete a planarly constrained task) to generate natural or semantic directions. This method utilizes the camera projection principle, which implies that a 2D axis $\mathbf{v}_c$ on the image plane may be projected from an infinite number of vectors $\mathbf{v}_d$ in 3D space. Robot movements along any of these 3D vectors cause the on-screen robot to move in the same direction.

Let $\mathbf{R}_c \in SO(3)$ denote the orientation of the camera with respect to the robot. The orientation of the meaningful plane is represented by a vector $\mathbf{v}_p \in \mathbb{R}^3$ that is perpendicular to the plane. We assume that $\mathbf{R}_c$ and $\mathbf{v}_p$ are known through some robot-camera calibration method (\textit{e.g.}, \cite{ilonen2011robust}) or geometric constraint inferring algorithm (\textit{e.g.}, \cite{subramani2018inferring}). Let $\mathbf{v}_c \in \mathbb{R}^2 $ be an axis in the camera image plane. The projected camera axis $\mathbf{x} \in \mathbb{R}^3$ can be computed using the following two equations.  
\begin{align}
     & \mathbf{P} \,\, \mathbf{R}_c^\intercal \mathbf{x} = \mathbf{v}_c \label{eq:projection1} \\
     & \mathbf{v}_p \cdot \mathbf{x} = 0 \label{eq:projection2}
\end{align}
where $\mathbf{P}=\begin{bmatrix}
1 & 0 & 0\\
0 & 1 & 0
\end{bmatrix}$ is a projection matrix that projects a vector onto the camera image (XY) plane.
Equation \ref{eq:projection1} makes sure that the 2D on-screen robot movements match user expectations to minimize visual-motor misalignments. Meanwhile, Equation \ref{eq:projection2} guarantees that the 3D robot movements $\mathbf{x}$ are within the natural or semantic plane. 
Combining Equations \ref{eq:projection1} and \ref{eq:projection2}, a projection camera axis can be obtained given a camera axis $\mathbf{v}_c$, camera orientation $\mathbf{R}_c$, a vector $\mathbf{v}_p$ perpendicular to the meaningful plane:
\begin{equation}
    \text{proj}(\mathbf{v}_c, \mathbf{R}_c, \mathbf{v}_p) = \mathbf{x} = \begin{bmatrix}
        \mathbf{P}\,\mathbf{R}_c^\intercal \\
        \mathbf{v}_p ^ \intercal
    \end{bmatrix}^{-1} \begin{bmatrix}
        \mathbf{v}_c \\
        0
    \end{bmatrix} \label{eq:projection}
\end{equation}

\begin{figure*} [tb]
    \centering
    \includegraphics[width=0.8\textwidth]{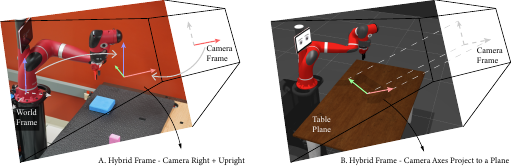}
    % \vspace{-0.3cm}
    \caption{\textbf{A.} A hybrid frame that is constructed by combining the camera right axis and the upright direction. \textbf{B.} A hybrid frame that is formulated by projecting camera axes onto a plane.}
    % \Description{new_mappings}
    % \vspace{-0.4cm}
    \label{fig: new_mappings}
     \vspace{-0.3cm}
\end{figure*}

Considering both the 3D robot movements before projection and the 2D on-screen robot movements after projection, the projected camera axes (Figure \ref{fig: new_mappings}B) minimize visual-motor misalignments as well as satisfy the natural or semantic requirements.
\subsection{Organizing Existing \ControlFrames{}} \label{sec:review}
With our design space, a \controlframe{} can be described by how an input device axis is mapped to the remote environment. 
Our design space provides a vocabulary to organize and differentiate \controlframes{}. The vocabulary is simple yet descriptive enough to represent various \controlframes{} and precise enough to maintain reproducibility. In order to demonstrate that our design space serves as a structure to organize and differentiate \controlframes{}, in this section, we systematically identify \controlframes{} in prior works and position them in our design space.

\begin{table*}[tb]
  \footnotesize
  \caption{ \Controlframes{} in prior works or designed in our case studies }
  \label{tab:mappings}
%    \begin{tabularx}{\textwidth}{XXXX}
\begin{tabularx}{\textwidth}{bmms}
    \Xhline{0.5\arrayrulewidth} 
    \multicolumn{4}{c}{\rule{0pt}{1.1\normalbaselineskip}%
    3D Input Device (\textit{e.g.}, VR Controller or Space Mouse)} \\[1mm]
    \Xhline{0.3\arrayrulewidth} 
    \rule{0pt}{1.1\normalbaselineskip}%    
    Name & $x$-axis$^1$ & $y$-axis$^2$ & $z$-axis$^3$ \\[1mm]
    \Xhline{0.3\arrayrulewidth}
    \rule{0pt}{1.1\normalbaselineskip}%
    Robot frame \cite{hiatt2006coordinate, rakita2018autonomous, quere2020shared, chotiprayanakul2009workspace, lamb2005human, praveena2022understanding} & Robot right & Robot forward & Robot up \\
    \rule{0pt}{1.1\normalbaselineskip}%
    Camera frame \cite{hiatt2006coordinate, draelos2017teleoperating, dejong2011mental} & Camera right & Camera forward & Camera up \\
    \rule{0pt}{1.1\normalbaselineskip}%
    End-effector frame  \cite{hiatt2006coordinate, draelos2017teleoperating, lamb2005human} & End-effector x-axis & End-effector y-axis & End-effector z-axis  \\
    \rule{0pt}{1.1\normalbaselineskip}%
    Orbit control - Fig. \ref{fig: existing_frames}A  \cite{abi2016visual} & End-effector right (in a spherical coordinate system) & End-effector up (in a spherical coordinate system) & End-effector back  \\
    \rule{0pt}{1.1\normalbaselineskip}%
    Pour task frame - Fig. \ref{fig: existing_frames}B  \cite{quere2020shared} & Bottle up (in a polar coordinate system)  & Bottle up (in a polar coordinate system) & Vertically up \\
    \rule{0pt}{1.1\normalbaselineskip}%
    Camera right + upright$^*$ - Fig. \ref{fig: new_mappings}A \cite{rakita2018autonomous, praveena2022understanding} & Camera right & World upright $\times$ camera right ** & World upright \\
    \Xhline{0.3\arrayrulewidth} 
\end{tabularx}
  
  \begin{tabularx}{\textwidth}{BMM}
    \multicolumn{3}{c}{\rule{0pt}{1.1\normalbaselineskip}%
    2D Input Device (\textit{e.g.}, Computer Mouse or Joystick)} \\[1mm]
    \Xhline{0.3\arrayrulewidth} 
    \rule{0pt}{1.1\normalbaselineskip}%
    Name & $x$-axis$^1$ & $y$-axis$^2$ \\[1mm]
    \Xhline{0.3\arrayrulewidth} 
    \rule{0pt}{1.1\normalbaselineskip}%
    Robot frame  \cite{talha2019preliminary, wu2020development} & Robot right & Robot forward   \\
    \rule{0pt}{1.1\normalbaselineskip}%
    Camera frame  \cite{baaberg2016adaptive, atherton2009supporting, wu2021camera, talha2019preliminary} & Camera right & Camera up \\
    \rule{0pt}{1.1\normalbaselineskip}%
    End-effector frame \cite{talha2019preliminary, wu2020development, nawab2007joystick, chintamani2009improved} & End-effector x-axis & End-effector y-axis \\
    \rule{0pt}{1.1\normalbaselineskip}%
    Orbit control - Fig. \ref{fig: existing_frames}A \cite{baaberg2016adaptive, talha2019preliminary}& End-effector right (in a spherical coordinate system)  & End-effector up (in a spherical coordinate system) \\
    \rule{0pt}{1.1\normalbaselineskip}%
    Spray task frame - Fig. \ref{fig: existing_frames}C \cite{mower2019comparing}& Surface right & Surface up \\
    \rule{0pt}{1.1\normalbaselineskip}%
    View-dependent frame - Fig. \ref{fig: existing_frames}D  \cite{notheis2014evaluation}& World axes that form acute angles with the camera image plane & World axes that form acute angles with the camera image plane\\
    \rule{0pt}{1.1\normalbaselineskip}%
    Learning-based mapping \cite{pierce2012data, li2020learning, losey2020controlling, mehta2022learning}  & A direction decided by a learned function & A direction decided by a learned function \\
    \rule{0pt}{1.1\normalbaselineskip}%
    Projected camera axes$^*$ - Fig. \ref{fig: new_mappings}B & Camera right projects to a plane & Camera up projects to a plane \\[1mm]
    %\Xhline{2\arrayrulewidth}
    \Xhline{0.3\arrayrulewidth} 
    \multicolumn{3}{l}{\rule{0pt}{1\normalbaselineskip}%
    1 Mapped from the input device's right axis} \\ 
    \multicolumn{3}{l}{2 Mapped from the input device's forward axis} \\
    \multicolumn{3}{l}{3 Mapped from the input device's up axis} \\
    \multicolumn{3}{l}{$^*$ Hybrid frames designed in our case studies in Section \ref{sec:case_studies}. } \\
    \multicolumn{3}{l}{$^{**}$ $A \times B$ denotes the cross product of the directions A and B.} 
  \end{tabularx}
\end{table*}

\begin{figure*} [tb]
    \centering
    \includegraphics[width=\textwidth]{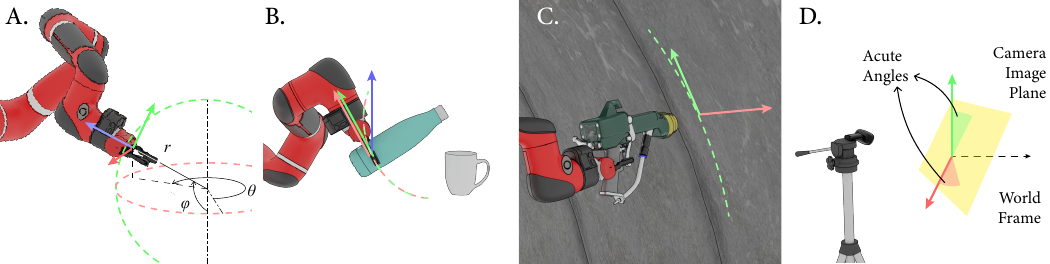}
    \caption{Visualization of some \controlframes{} in prior works. Red, green, and blue axes represent the directions mapped from the input device right (x-axis), forward (y-axis), and up (z-axis), respectively. \Controlframes{} A and B are designed for 3D input devices; C and D are for 2D input devices. \textbf{A. Orbit control} \cite{abi2016visual, baaberg2016adaptive, talha2019preliminary} - The \controlframe{} moves on an imaginary sphere, whose radius is controlled by the blue axis. \textbf{B. Pour task frame} \cite{quere2020shared} - Vertical inputs move the bottle vertically up and down (blue arrow). Inputs along other two axes tilt the bottle (red and green) to finish the pouring task. \textbf{C. Spray task frame} \cite{mower2019comparing} - 2D inputs are mapped to the surface to be sprayed. The figure shows a tunnel surface that is curved along the vertical direction. \textbf{D. View-dependent frame} \cite{notheis2014evaluation} - The \controlframe{} is a plane formed by two axes chosen from the world frame. The two axes must form an acute angle with the camera image plane.}. 
    \label{fig: existing_frames}
     \vspace{-0.3cm}
\end{figure*}

\emph{Methodology} - We collected articles on Google Scholar by combining keywords in [teleoperation, telemanipulation, robot] and [input mapping, control mapping, control frame, reference frame, coordinate system].  
Since all telemanipulation systems involve a choice of \controlframes{}, we only included works that either introduce a new coordinate system or compare multiple coordinate systems. A total of 18 papers were identified.

\emph{Findings} - Our findings are listed in Table \ref{tab:mappings}. All \controlframes{} we found can be represented in our design space. Many \controlframes{} align with key coordinate frames (\textit{e.g.}, the end-effector frame) in the remote environment.
Some of the \controlframe are attached to some imaginary objects. For example, several works  \cite{talha2019preliminary, baaberg2016adaptive, abi2016visual} drew inspiration from orbit camera control methods in computer graphics. Orbit control moves the end-effector along the surface of an imaginary sphere, creating orbital motions (Figure \ref{fig: existing_frames}A).
% Task-specific
Quere \textit{et al.} \cite{quere2020shared} and Mower \textit{et al.} \cite{mower2019comparing} use task-specific \controlframes{} that utilize semantic directions (as shown in Figure \ref{fig: existing_frames} B\&C).  
Finally, some prior works \cite{notheis2014evaluation, rakita2018autonomous} use hybrid frames. As an example, the view-dependent frame presented by Notheis \textit{et al.} \cite{notheis2014evaluation} (Figure \ref{fig: existing_frames}D) consists of the world axes that form the closest acute angles with the camera image plane. 
By positioning the existing \controlframes{} in our design space, we demonstrate that the design space serves as a structure to organize and differentiate \controlframes{}.  

\subsection{Design Criteria} \label{sec:criteria}
With so many possible options in our design space, designers need a method to reason about and predict the suitability of \controlframes{} for a scenario. 
We provide three initial design criteria based on prior work and our design experience. 
While designers should consider all the criteria comprehensively, we note that there are trade-offs between the criteria and it may be impossible to satisfy all the design criteria simultaneously in some scenarios.

Criterion  1: \emph{Minimize Visual-Motor Misalignments} --- We define visual-motor misalignments as the angular difference between the control coordinate system and the camera frame. Such angular difference causes a misalignment between the robot motion on screen (visual) and the user's input motion (motor), increasing the human operator's mental workload \cite{dejong2011mental} and affect teleoperation performance \cite{dejong2004improving, rakita2018autonomous}. However, the detrimental effects of visual-motor misalignments are isotropic \cite{ellis2012human} and can fluctuate based on operators' spatial abilities \cite{menchaca2007influence}. Visual-motor misalignments can be described using rotations about three axes: \emph{roll} axis pointing to the operator's straight ahead, \emph{pitch} axis to the operator's left, and \emph{yaw} axis directed upward. Ellis \textit{et al.} \cite{ellis2012human} found that the rotation along pitch or yaw axes was distinctly less disruptive than roll rotation. Such findings are consistent with the usage of a computer mouse - operators move a mouse in a horizontal table plane (motor) and see the cursor move in a nearly vertical screen plane (visual). Most operators can quickly adapt to this visual-motor misalignment which is approximately a 90-degree pitch rotation. In addition, the experiment conducted by Menchaca-Brandan \cite{menchaca2007influence} suggests that operators with better spatial abilities can better handle visual-motor misalignments. The operator's ability to adapt to visual-motor misalignments gives designers more creative freedom when designing \controlframes{}, allowing them to take the following two design criteria into consideration.

Criterion  2: \emph{Be Natural} --- Natural \controlframes{} enable natural mapping, taking advantage of spatial analogies and leading to immediate understanding \cite{norman2013design}. For example, to move the robot up, move the controller up. To rotate the end-effector about the vertical axis, rotate the controller in the same way.
From the perception point of view, humans inherently maintain a stable perception of the vertical direction despite changing viewpoints, which is known as \emph{orientation constancy} \cite{kheradmand2017perception}. Due to orientation constancy, even if the camera is tilted in the remote environment, a human operator still can perceive whether the end-effector moves vertically or not.
Natural mappings work the way that the user expects them to, without having to think about it \cite{campeau2018intuitive}.  
When human operators move horizontally, they expect the robot also to move parallel to the ground plane. Therefore, a natural \controlframe{} for horizontal 2D input devices (\textit{e.g.}, computer mice) moves the robot horizontally.

Criterion  3: \emph{Consider Task Semantics} --- While the first two criteria focus on mental workload and intuitiveness for human operators, the third criterion is proposed from the task point of view. Designers should consider what robot motions are desired to complete the task and how to design the \controlframe{} to facilitate the operations that enable those desired robot motions. In particular, \controlframes{} should be chosen to match the semantic directions for those geometrically constrained tasks. For example, Mower \textit{et al.} \cite{mower2019comparing} present a robotic system in which a 2D joystick is used to teleoperate a robot to spray concrete onto a curved tunnel surface. In this task, desired motions maintain a fixed distance to the tunnel surface, so they use a control frame containing semantic directions that are parallel to the surface (Figure \ref{fig: existing_frames}C). The control coordinate system facilitates teleoperation by consistently keeping the nozzle at a constant distance from the surface. 

When task semantics are taken into consideration, the designed \controlframes{} are task-specific and may not be applicable or meaningful for other tasks. There is a trade-off: task-specific control frames may greatly facilitate the teleoperation of a particular task; however, when performing other tasks, the system needs to switch control frames, imposing an additional cognitive load for operators to adapt to the new frames. The control frames for some special-purpose teleoperation systems can be task-specific, while general-purpose robots should use universal \controlframes{}.

\subsection{Summary}
A \controlframe{} can be designed by selecting a direction in the remote environment for each axis of the input device. 
We enumerated common choices of these directions, including the axes of key coordinate frames, semantic directions, and projected camera axes. 
Moreover, we provided design criteria that suggest designers to consider visual-motor misalignments, naturalness, and task semantics when designing \controlframes{}.

\section{Case Studies} \label{sec:case_studies}

In the previous section, we demonstrated the \textit{descriptive} power of our design space by organizing existing \controlframes{} in the literature. In this section, we demonstrate its \textit{generative} and \textit{evaluative} power using three case studies. In each case study, we consider a telemanipulation scenario and use the design space to \emph{generate} an \controlframe{} and make predictions about its effectiveness over some common designs (\textit{i.e.}, \textit{evaluate}). We confirm these predictions with human subjects experiments in Section \ref{sec:evaluation}. 
The scenarios include a pick-and-place task and a geometrically constrained tracing task and use a 6 DoF VR controller or a 2D mouse as input devices (Figure \ref{fig: study_design}). 
The designed \controlframes{} are examples of \emph{hybrid frames}, which leverage geometric information of different coordinate frames. 

\subsection {Pick-and-Place with VR Controller} \label{sec:case_studies:inperson}
In this scenario, operators perform a pick-and-place task (Figure \ref{fig: study_design} Row 1) with a 3D input device. 
Common choices of \controlframes{} are the \emph{robot frame} or \emph{camera frame}.
Choosing the \emph{robot frame} as the \controlframe{ }would cause significant visual-motor misalignments (Section \ref{sec:criteria} Criterion 1) because the camera and the robot are facing different directions, leading to a large angular difference between the robot frame and the camera frame. Meanwhile, the \emph{camera frame} lacks naturalness (Section \ref{sec:criteria} Criterion 2) because when an operator moves the controller vertically up, the robot moves along the camera up direction, which is not vertical. 

With our design space, \controlframes{} are designed by picking a direction in the remote environment for each axis of the input device.  
Firstly, we picked the right direction of the camera for the input device's right. The camera right direction partially encodes camera orientation and reduces visual-motor misalignments. When an operator moves right, they see the end-effector moving right on-screen. Secondly, we chose to map the input device's up axis to the world up direction, which is a natural direction that aligns with gravity. To move the end-effector straight up, an operator just moves the controller straight up. Finally, we generated the third mapping direction by taking the cross-product of the first two directions. We made this design choice to get a right-handed coordinate frame.
The camera right direction, world up direction, and the cross product of these two directions form \emph{hybrid frame 1} in the remote environment (Figure \ref{fig: new_mappings}A).
The mathematical formulation of the hybrid frame is in Table \ref{tab:input_mapping_math}. 

We predicted that \emph{hybrid frame 1} would lead to better task performance and user experience than the \emph{robot} or \emph{camera frame} because the hybrid frame considers both visual-motor alignments and naturalness (Section \ref{sec:criteria} Criteria 1\&2). 
Although \emph{hybrid frame 1} is similar to the camera frame with only some rotational difference along the camera pitch axis (see the visualization in Figure \ref{fig: study_design} Row 1), we predicted that the subtle difference would improve user performance. 

\subsection {Pick-and-Place with Mouse} \label{sec:case_studies:turk1}
In the second case study, operators perform a pick-and-place task with a mouse (Figure \ref{fig: study_design} Row 2). As described in \cref{sec:related_work}, prior research has found that low-dimensional input devices, including a computer mouse, are more accurate, comfortable, and easier to use for inexperienced operators over 3D input devices. 
Since a mouse only captures user movement along a 2D plane, the mouse wheel is utilized to provide an additional dimension to perform the pick-and-place task in 3D space. 
People habituate the mapping between a mouse and its cursor on a screen, \textit{e.g.}, to move the cursor up, move the mouse forward. To replicate a similar mapping in telemanipulation, a system selects the \emph{camera frame} as the control frame, which makes on-screen end-effector movements appear similar to a cursor.
However, the \emph{camera frame} lacks naturalness (Section \ref{sec:criteria} Criterion 2) because operators rarely want the end-effector to move parallel to the camera image plane. Considering naturalness, the robot should move parallel to the table plane because operators move the mouse in the horizontal table plane. 
Another common design choice maps inputs to the \emph{robot frame}. However, given the camera position, a mapping to the robot frame would cause significant visual-motor misalignments (Section \ref{sec:criteria} Criterion 1). 

With our design space, we designed a \controlframe{} with \emph{projected camera axes}. As described in Section \ref{sec:choices}, the projected camera axes utilize the camera projection principle, allowing on-screen movements to minimize visual-motor misalignments and 3D robot motions to be natural (Criteria 1\&2 in Section \ref{sec:criteria}). To create a right-handed coordinate frame, the mouse wheel was selected to move the end-effector along the direction that is orthogonal to the projected camera axes. The projected camera axes and the direction orthogonal to them form \emph{hybrid frame 2}, whose mathematical formulation is in Table \ref{tab:input_mapping_math}. Considering both visual-motor misalignment and naturalness, we predicted that \emph{hybrid frame 2} would improve performance and user experience compared to the \emph{robot} or \emph{camera frame}.

\subsection {Tracing with Mouse}  \label{sec:case_studies:turk2}
\begin{figure*} [tb]
    \centering
    \includegraphics[width=\textwidth]{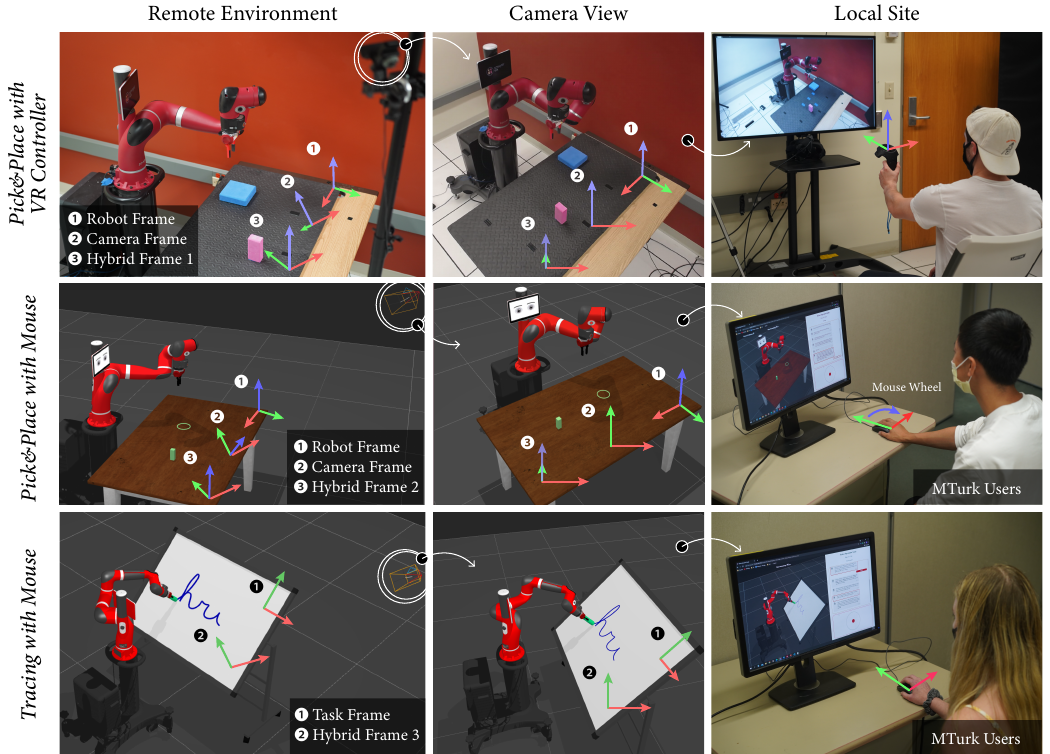}
    \caption{We conducted three case studies in which we applied the choices and reasoning afforded by the design space to design \controlframes{} for various telemanipulation scenarios. The \controlframe{} designed in each case study was assessed in an independent human subject experiment. The designed \emph{hybrid frames} were experimental conditions, and the \emph{robot frame}, \emph{camera frame}, and \emph{task frame} were selected as control conditions. In the visualization, red, green, and blue arrows represent the directions mapped from the input device's right, forward and up axes, respectively.}
    % \Description{hybrid mappings}
    \label{fig: study_design}
\end{figure*}

In the third case study, operators trace letters on a whiteboard with a mouse; this is a planarly-constrained task where the pen tip should stay on the 2D whiteboard plane (Figure \ref{fig: study_design} Row 3). One common design choice of \controlframes{} is the whiteboard's coordinate frame (\emph{task frame}), but it will cause visual-motor misalignments (Section \ref{sec:criteria} Criterion 1) since the camera is not facing the whiteboard. 

In the \controlframe{} that we designed for this scenario, we mapped mouse inputs to the camera axes projected on the whiteboard plane (\emph{hybrid frame 3}). This decision was made by following Criteria 1 and 3 in Section \ref{sec:criteria}:minimize visual-motor alignments and consider task semantics. The \controlframe{} is semantic because it makes the end-effector stay in the whiteboard plane. The \controlframe{} minimizes visual-motor misalignment because when being watched through a camera, the end-effector movement on screen behaves like a mouse cursor (\textit{e.g.}, when an operator moves the mouse to the right, the end-effector on screen moves right like a mouse cursor). Therefore, we predicted that \emph{hybrid frame 3} would outperform the \emph{task frame}. Table \ref{tab:input_mapping_math} lists the mathematical formulation to implement \emph{hybrid frame 3}.

\section{Evaluation} \label{sec:evaluation}

In the previous section, we designed \controlframes{} for three telemanipulation scenarios and reasoned about their suitability with our design space. 
In this section, we evaluate the designed \controlframes{} to explore whether the choices and reasoning afforded by our design space allowed us to predict appropriate \controlframes{} for various scenarios.
Our hypothesis is that the predictions made using the design space are accurate. Specifically, our design space predicts that the hybrid frames designed with our design space will improve performance and user experience compared to some commonly-used reference frames.

\subsection{Experimental Design, Tasks, \& Conditions }

In the case studies in Section \ref{sec:case_studies}, we designed \controlframes{} for three scenarios by picking meaningful directions and following the design criteria in our design space.
The \controlframe{} designed for each scenario was evaluated in an independent experiment with different participants. Each experiment followed a within-participants design.
The order of the conditions in each experiment was counterbalanced. Experiment 1 was conducted in-person using a physical robot, while experiments 2 and 3 were conducted online using a simulation platform with participants recruited via Amazon Mechanical Turk. In all three of our experiments, cameras were placed in surveillance-camera-like positions to provide adequate coverage of the workspace and non-occluded views from an elevated perspective. Figure \ref{fig: study_design} visualizes the \controlframes{} assessed in the three independent experiments. The mathematical representations of the axes of each \controlframes{} are listed in Table \ref{tab:input_mapping_math}.

\begin{table*}[tb]
  % \helvetica
  \footnotesize
  \caption{\Controlframes{} evaluated in our experiments}
  \vspace{-2mm}
  \label{tab:input_mapping_math}
  \begin{tabular}{p{2.5cm}p{3.5cm}p{3.8cm}p{3.8cm}p{1.4cm}}
    \Xhline{0.5\arrayrulewidth}
    \rule{0pt}{1.1\normalbaselineskip}%
    Experiment & \ControlFrame{} & 
    $\mathbf{d}_x$ in Eq. \ref{eq: mapping_matrix} & 
    $\mathbf{d}_y$ in Eq. \ref{eq: mapping_matrix} & 
    $\mathbf{d}_z$ in Eq. \ref{eq: mapping_matrix} \\[1mm]
    \hline
    \rule{0pt}{1.1\normalbaselineskip}%    
    \multirow{3}{2.3cm}{E1 - Pick \& Place \\ with VR Controller} & Robot Frame & $[1,0,0]^\intercal$ & $[0,1,0]^\intercal$ & $[0,0,1]^\intercal$ \\
    & Camera Frame & $\mathbf{R}_c [1,0,0]^\intercal$ & $\mathbf{R}_c [0,0,-1]^\intercal$ & $\mathbf{R}_c [0,1,0]^\intercal$ \\
    & Hybrid Frame 1 & $\mathbf{R}_c [1,0,0]^\intercal$ & $\mathbf{R}_w [0,0,1]^\intercal \times \mathbf{R}_c [1,0,0]^\intercal$ & $\mathbf{R}_w [0,0,1]^\intercal$ \\[1mm]
    \hline
    \rule{0pt}{1.1\normalbaselineskip}%    
    \multirow{3}{2.3cm}{E2 - Pick \& Place \\ with Mouse} & Robot Frame & $[1,0,0]^\intercal$ & $[0,1,0]^\intercal$ & $[0,0,1]^\intercal$ \\
    & Camera Frame & $\mathbf{R}_c [1,0,0]^\intercal$ & $\mathbf{R}_c [0,1,0]^\intercal$ & $\mathbf{R}_c [0,0,1]^\intercal$ \\
    & Hybrid Frame 2 & $\text{proj}([1,0]^\intercal, \mathbf{R}_c, \mathbf{R}_w[0,0,1]^\intercal)$ & $\text{proj}([0,1]^\intercal, \mathbf{R}_c,  \mathbf{R}_w[0,0,1]^\intercal)$ & $\mathbf{R}_w [0,0,1]^\intercal$ \\[1mm]
    \hline
    \rule{0pt}{1.1\normalbaselineskip}%    
    \multirow{2}{2.3cm}{\makecell[l]{E3 - Tracing \\ with Mouse}
    } & Task Frame & $\mathbf{R}_t [1,0,0]^\intercal$ & $\mathbf{R}_t [0,1,0]^\intercal$ & \multicolumn{1}{c}{/} \\
    & Hybrid Frame 3 & $\text{proj}([1,0]^\intercal, \mathbf{R}_c, \mathbf{R}_t[0,0,1]^\intercal)$ & $\text{proj}([0,1]^\intercal, \mathbf{R}_c,  \mathbf{R}_t[0,0,1]^\intercal)$ & \multicolumn{1}{c}{/} \\[1mm]
    \Xhline{\arrayrulewidth}
    \multicolumn{5}{p{6.8in}}{\rule{0pt}{1\normalbaselineskip}A \controlframe{} is obtained by substituting $\mathbf{d}_x, \mathbf{d}_y, \mathbf{d}_z$ into Equation \ref{eq: linear_transform}. The orientation of the camera, the world coordinate system, and the task frame, $\mathbf{R}_c, \mathbf{R}_w, \mathbf{R}_t \in SO(3)$, are all with respect to the robot. Projected camera axis $\text{proj}(\mathbf{v}_c, \mathbf{R}_c, \mathbf{v}_p)$ is obtained using Equation \ref{eq:projection}. } 
  \end{tabular} 
\end{table*}

\subsubsection{Experiment 1: Pick-and-Place with VR Controller} Participants picked up a foam block and placed it onto another flat foam block. Participants were instructed to move as quickly as possible but avoid colliding with the table or knocking over the foam block. To simplify the experiment, the orientation of the end-effector was set to grasp the foam block and only translational inputs were mapped to the robot. The robot was controlled with one of the three \controlframes{} described in Section \ref{sec:case_studies:inperson}: the \emph{robot frame} and \emph{camera frame} as control conditions; \emph{hybrid frame 1} (camera right + upright) as the experimental condition. 

\subsubsection{Experiment 2: Pick-and-Place with Mouse} Participants picked up a block and placed it in a given circular area. Participants were instructed to move as quickly as possible while avoiding collisions. 
The translational movements of the mouse were mapped to the robot.
As described in Section \ref{sec:case_studies:turk1}, the \emph{robot frame} and \emph{camera frame} were employed as control conditions, and \emph{hybrid frame 2} (camera axes project to the table plane) was the experimental condition. 

\subsubsection{Experiment 3: Tracing with Mouse} Participants were instructed to trace the letters on a whiteboard as quickly as possible while maintaining accuracy. As described in Section \ref{sec:case_studies:turk2}, the experimental condition mapped user inputs to the \emph{hybrid frame 3} (camera axes project to the whiteboard plane). The control condition was the \emph{task frame}, which mapped user inputs to the whiteboard's coordinate frame. Both conditions guaranteed that the pen tip always stayed on the whiteboard.

\subsection{Implementation Details}

We implemented a teleoperation system for the in-person experiment 1 and a web-based platform for the online experiments 2 and 3. Performing the experiments online provides access to a larger and more diverse participant pool.

In experiment 1 (in-person pick and place), participants used a mimicry-based telemanipulation approach \cite{rakita2020effects, rakita2018relaxedik} to guide the 3D position of the end-effector of a Rethink Sawyer robot.
Participants were located in the same room with the robot and a room divider was placed to obstruct the participants' line of sight. The participants' inputs $[\Delta x, \Delta y, \Delta z]^\intercal$ were captured by an HTC Vive motion controller at approximately 60 Hz. The velocity of the end-effector was computed using Equation \ref{eq: linear_transform}, where $\mathbf{v}_i=[\Delta x/\Delta t, \Delta y/\Delta t, \Delta z/\Delta t]^\intercal$ and $\boldsymbol{\omega}_i=[0,0,0]^\intercal$.
End-effector positions were filtered by a collision protection program, which halted the control if the robot's fingertip was closer than 1 centimeter to the tabletop. 
Participants were not aware of the collision protection program, but they were instructed to avoid collisions with the table.
The trigger on the Vive controller was used to open or close the robot's gripper. We used a Logitech 930e webcam to stream 1080P video on a large-screen display.

For experiments 2 and 3, we used an online study platform that allowed multiple participants to teleoperate separated virtual robots simultaneously. We ran the Robot Operating System (ROS) on a Google Cloud Platform server to provide a forward/inverse kinematics solver and store teleoperation data. Each time a client connected to the ROS server, a new virtual robot and a RelaxedIK \cite{rakita2018relaxedik} inverse-kinematics solver were spawned for the client. The robot link positions were sent to the client, and the robot was finally rendered using \texttt{three.js} in the client's browser.

To ensure our online study interface could run adequately on a participant's machine and the participant's internet condition was fast and stable enough for real-time telemanipulation, the participant must pass a technical qualification test before starting the study. The technical qualification page allowed users to compare side-by-side the rendered robot with a static image that showed a correctly rendered robot. Once the participant confirmed that the robot was rendered correctly, 300 commands were automatically sent to the server at 30 Hz to move the virtual robot along a predefined trajectory. During the automatic manipulation, latency was recorded and the study could only proceed if the two-way latency was always below 125 ms. This threshold was set based on prior research which found that teleoperation performance does not degrade until 250 ms of latency \cite{rakita2020effects, lum2008quantitative}. After the technical qualification test, a preliminary page ensured that the participant used a desktop or laptop computer (instead of a tablet) and a physical mouse (instead of a touchpad). Since people may have different scrolling direction settings (natural or reverse scrolling), we also calibrated scrolling directions by instructing participants to scroll mouse wheels towards or away from them.

Upon passing the technical qualification and preliminary test, participants controlled a virtual Rethink Sawyer robot using a mouse. By clicking in the starting area, participants started the control and their cursors were temporally locked. 
Participants picked up their mice for clutching and pressed the ESC button on keyboards to stop the control. 
In addition to the mouse movement on tabletop $[\Delta x, \Delta y]$, mouse wheels provided an additional input $\Delta z$ to perform the 3D pick-and-place task. The robot speed was computed using Equation \ref{eq: linear_transform}, where $\mathbf{v}_i=[\Delta x/\Delta t, \Delta y/\Delta t, \Delta z/\Delta t]^\intercal$ and $\boldsymbol{\omega}_i=[0,0,0]^\intercal$.
In the pick-and-place task, a virtual block was automatically grasped if it was close enough to the gripper. A warning sound was played if the robot's gripper or the object in the gripper collided with the table.

\subsection{Experimental Procedure}
All three of our experiments followed the same procedure. After informed consent, participants were provided with study goals and instructed on using the input devices to control the robot.
Then participants were presented the first condition. After a fixed amount of time of practice to get used to the \controlframe{} (1 minute for the in-person experiment 1 and 30 seconds for online experiments 2 and 3), 
participants were introduced to the study task and finished two practice rounds.
Then participants performed the task for three trials and filled out a questionnaire regarding their experience in the condition. This procedure was repeated for other conditions.

In both pick-and-place experiments, the block's initial and target positions change in each practice round and trial. For the tracing task, participants traced a four-point polyline and a semicircular arc as practice and traced letters for each condition in the order ``hri'', ``ros'', and ``lab''. 
Upon finishing all conditions, participants provided demographic information, and 
% answered two questions "which condition do you prefer?" and "please briefly describe the differences between the robots?". 
were asked their preferred condition and to describe the differences between the robots.
The two questions were asked in a semi-structured interview in the in-person experiment and in a Qualtrics questionnaire in online experiments.  
The in-person pick-and-place experiment took approximately 40 minutes, and participants received \$10 compensation. Online participants received \$3 for about 20 minutes in the tracing experiment or \$5 for about 30 minutes for finishing the pick-and-place experiment.

\subsection{Participants}

We recruited 29 participants on campus for experiment 1 (pick \& place with VR controller). Five participants were excluded from the analyses because of failing to complete all three trials in one condition. The resulting 24 participants (1 demifemale, 8 females, and 15 males) aged 18 to 32. They had various education backgrounds, including engineering, psychology, history, economics, physics, and management. 
We recruited participants via Amazon Mechanical Turk for our online experiments. 
% Pickplace
The 24 participants for experiment 2 (online pick \& place) aged 24 to 68 (7 females and 17 males). They had various occupations, including electrician, IT manager, baker, sales associate, and legal transcriptionist.
% Tracing
In online experiment 3 (tracing with mouse), one participant was excluded from the analyses for leaving the drawing area (the whiteboard). The remaining 24 participants (1 agender, 10 females, and 13 males) aged 28 to 57. They were from various industries, including information technology, education, food service, and hospitality.

Participants reported their familiarity with relevant technologies using 5-point Likert scales. As shown in Table \ref{tab:demographics}, all participants had low-to-moderate familiarity with robots and Computer-Aided Design (CAD) software. As for the familiarity with 3D video games, in-person participants had moderate familiarity, while online participants had moderate-to-high familiarity. We speculate the difference was caused by the technical qualification test before online studies. To pass the technical qualification test, participants required a fast computer and internet, and these types of participants tended to be game players. 

\begin{table}[tb]
  \footnotesize
  \center
  \caption{Participant demographics}
  \label{tab:demographics}
  \vspace{2mm}
  \resizebox{\columnwidth}{!}{%
  \begin{tabular}{ccccc}
    \Xhline{0.5\arrayrulewidth}
    \rule{0pt}{1.1\normalbaselineskip}% 
    Experiment & Age & \multicolumn{3}{c}{ Familiarity with } \\
    & & Robots &  3D Video Games &  CAD Software \\
    \Xhline{0.5\arrayrulewidth}
    \makecell[l]{
    \rule{0pt}{1.1\normalbaselineskip}%
    E1 - Pick \& Place \\ with VR Controller} & \makecell[l]{$M=20.84$ \\ ($SD = 3.40$)} & \makecell[l]{$M=2.33$ \\ ($SD = 1.24$)} & \makecell[l]{$M=3.04$ \\ ($SD = 1.57$)} & \makecell[l]{$M=2.50$ \\ ($SD = 1.50$)} \\
    \Xhline{0.5\arrayrulewidth}
    \makecell[l]{
    \rule{0pt}{1.1\normalbaselineskip}%
    E2 - Pick \& Place \\ with Mouse} & \makecell[l]{$M=37.92$ \\ ($SD = 12.32$)} & \makecell[l]{$M=2.46$ \\ ($SD = 0.83$)} & \makecell[l]{$M=3.96$ \\ ($SD = 1.43$)} & \makecell[l]{$M=2.21$ \\ ($SD = 1.06$)} \\
    \Xhline{0.5\arrayrulewidth}
    \makecell[l]{
    \rule{0pt}{1.1\normalbaselineskip}%
    E3 - Tracing \\ with Mouse} & \makecell[l]{$M=38.67$ \\ ($SD = 7.72$)} & \makecell[l]{$M=2.50$ \\ ($SD = 0.93$)} & \makecell[l]{$M=3.63$ \\ ($SD = 1.35$)} & \makecell[l]{$M=2.00$ \\ ($SD = 0.98$)} \\
    \Xhline{0.5\arrayrulewidth}
  \end{tabular} }
\end{table}

\begin{figure*} [tb]
  \centering
  \includegraphics[width=\textwidth]{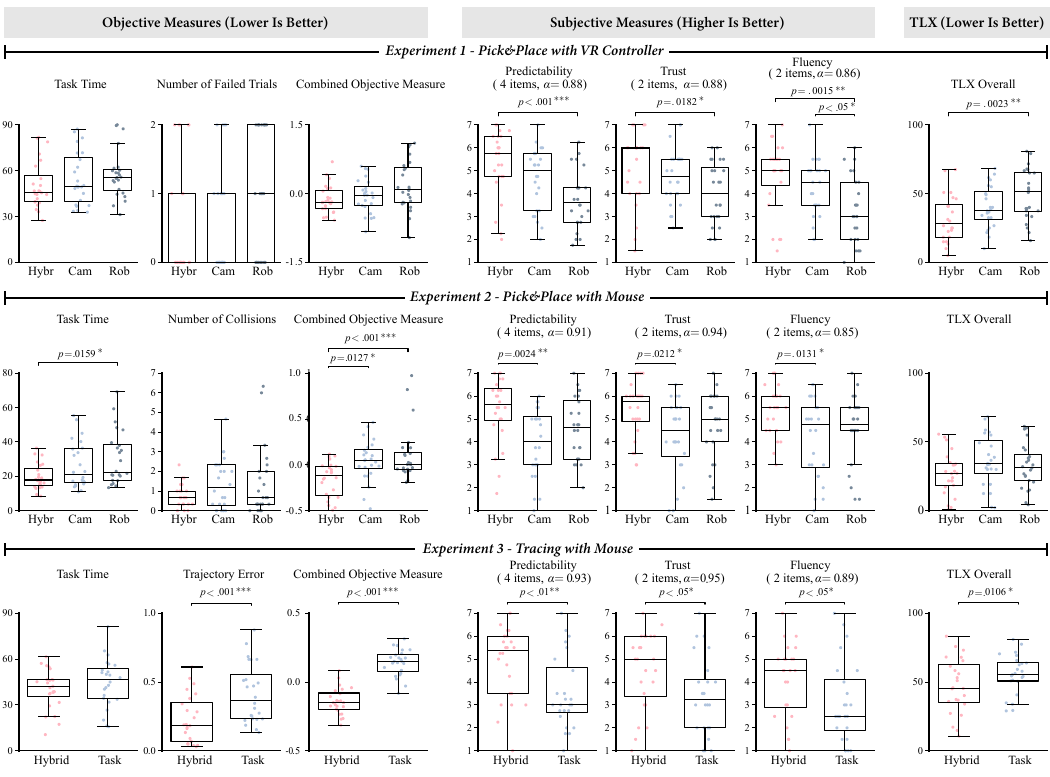}
  \caption{Box and whisker plots of data from the performance and user perception measures for each experiment. 
  The top and bottom of each box represent the first and third quartiles,
and the line inside each box is the statistical median of the data. The length of the box is defined as the interquartile range
(IQR). The whiskers are with maximum 1.5 IQR.
    Horizontal lines indicate significant Tukey HSD test results.
    TLX means NASA Task Load Index.
    }
  % \Description{results}
  \label{fig: results}
\end{figure*}

\begin{table}[tb]
  \center
 % \helvetica
  \footnotesize
  \caption{Statistical results of our measures}
  \label{tab: statistical_results}
  \vspace{2mm}
  \resizebox{\columnwidth}{!}{%
  \begin{tabular}{lllllll}
    \Xhline{0.5\arrayrulewidth}
    \multicolumn{7}{c}{
    \rule{0pt}{1.1\normalbaselineskip}%
    Experiment 1: Pick \& Place with VR Controller} \\[1mm]
     & Hybrid Frame & Camera Frame  & Robot Frame & &\\
    Measures & Mean (SD) & Mean (SD) & Mean (SD) & $F(2,46)$ & \quad $p$ & $\eta^2$ \\[1mm]
    \hline
    \rule{0pt}{1.1\normalbaselineskip}%
    Time & \textbf{50.47} (15.42) & 54.71 (17.29) & 57.46 (15.53) & 1.56 & $.$2216 & 0.06\\
    Num Failures & \textbf{0.58} (0.83) & \textbf{0.58} (0.78) & 0.96 (0.91) & 1.78 & $.$1802 & 0.07\\
    Combined Metric & \textbf{-0.12} (0.31) & 0.05 (0.37) & 0.18 (0.54) & 2.26 & $.$1161 & 0.09\\
    Predictability & \textbf{5.31} (1.45) & 4.64 (1.37) & 3.74 (1.26) & 8.87 & $<.$001 & 0.28\\
    Trust & \textbf{5.13} (1.59) & 4.75 (1.22) & 4.00 (1.35) & 5.59 & $<.$01 & 0.20\\
    Fluency & \textbf{4.77} (1.53) & 4.27 (1.25) & 3.29 (1.44) & 8.35 & $<.$001 & 0.27\\
    TLX Overall & \textbf{31.96} (17.03) & 40.12 (15.50) & 49.25 (18.62) & 11.08 & $<.$001 & 0.33\\[1mm]
    \Xhline{0.5\arrayrulewidth}
  %\end{tabular} }
  %\resizebox{0.95\columnwidth}{!}{
  %\begin{tabular}{llllll}
    \multicolumn{7}{c}{
    \rule{0pt}{1.1\normalbaselineskip}%
    Experiment 2: Pick \& Place with Mouse} \\[1mm]
    \Xhline{0.5\arrayrulewidth}
    \rule{0pt}{1.1\normalbaselineskip}%
    Time & \textbf{19.70} (7.44) & 26.36 (13.37) & 30.11 (15.73) & 10.71 & $<.$001 & 0.32\\
    Num Collisions & \textbf{0.78} (0.58) & \ 1.36 (1.28) & 1.38 (1.75) & 2.05 & $.$1400 & 0.08\\
    Combined Metric & \textbf{-0.16} (0.19) & \ 0.05 (0.22) & 0.11 (0.29) & 5.40 & $<.$01 & 0.19\\
    Predictability & \textbf{5.40} (1.43) & 3.95 (1.51) & 4.56 (1.37) & 6.24 & $<.$01 & 0.21\\
    Trust & \textbf{5.38} (1.25) & 4.23 (1.62) & 4.73 (1.45) & 3.92 & $<.$05 & 0.15\\
    Fluency & \textbf{5.40} (1.10) & 4.17 (1.72) & 4.58 (1.49) & 5.09 & $<.$05 & 0.18\\
    TLX Overall & \textbf{26.86} (15.39) & 36.09 (18.81) & 32.38 (16.21) & 7.46 & $<.$01 & 0.24\\
    \Xhline{0.5\arrayrulewidth}
    \multicolumn{7}{c}{
    \rule{0pt}{1.1\normalbaselineskip}%
    Experiment 3: Tracing with Mouse} \\[1mm]
  \end{tabular}}
  \resizebox{\columnwidth}{!}{
  \begin{tabular}{llllll}
     & Hybrid Frame &  Task Frame \\
    Measures \quad\quad\quad \quad   & Mean (SD) \quad\quad\quad  & Mean (SD) \quad\quad\quad   &  $F(1,23)$ \quad\quad   & \quad $p$  \quad \quad &  $\eta^2$ \\
    \Xhline{0.5\arrayrulewidth}
    \rule{0pt}{1.1\normalbaselineskip}%
    Time & \textbf{40.22} (12.92) & 44.46 (15.91) & 3.12 & $.$0907 & 0.12\\
    Trajectory Error & \textbf{0.23} (0.17) & 0.42 (0.22) & 17.75 & $<.$001 & 0.44\\
    Combined Metric & \textbf{-0.14} (0.10) & 0.14 (0.10) & 51.71 & $<.$001 & 0.69\\
    Predictability & \textbf{4.86} (1.58) & 3.53 (1.60) & 8.01 & $<.$01 & 0.26\\
    Trust & \textbf{4.50} (1.68) & 3.40 (1.71) & 5.08 & $<.$05 & 0.18\\
    Fluency & \textbf{4.08} (1.61) & 3.08 (1.78) & 4.76 & $<.$05 & 0.17\\
    TLX Overall & \textbf{47.08} (19.97) & 55.68 (14.28) & 7.75 & $<.$05 & 0.25\\
    \Xhline{0.5\arrayrulewidth}
  \end{tabular}}
\end{table}

\subsection{Measures} \label{sec:measures}
We employed a combination of objective and subjective measures to assess participants’ performance and user experience.

\subsubsection{Objective Measures}
We employed completion time and task error metrics to assess task performance in each experiment. The maximum time limit for each task is 90 seconds. 
In addition to task time, we counted the \emph{number of failed trails} as the task error metric in experiment 1 (in-person pick \& place). A trial is considered to be a failure if the participant exceeded the time limit, knocked over the foam block, or triggered the robot's collision protection program. 
In experiment 2 (online pick \& place), we measured the \emph{number of collisions} between the table and either the robotic gripper or the block in hand. 
In experiment 3 (online tracing task), we adapted a \emph{trajectory error} metric from prior work \cite{rakita2020effects} to assess how well participants trace a target curve. The trajectory error metric is the sum of a Cartesian accuracy score and a completeness score. The Cartesian accuracy score is the average error distance between the pen tip and its closest point on the target curve. The Cartesian accuracy is normalized to $[0,1]$. To compute the completeness score, we first associate an arc-length parameter value in $[0,1]$ to all points on the target curve. The completeness score of each point on the pen tip curve is the arc-length parameter value of its closest point on the target curve. We then use the maximum completeness score on points as the completeness score of the entire curve. The range of \emph{trajectory error} is $[0,2]$, where a lower value indicates a better trajectory. 

The time and task error metrics are possibly associated, \textit{e.g.}, a participant who hurriedly finished the pick-and-place task in a short time might have a large number of collisions. Therefore, we formulated a \emph{combined objective measure} for each experiment to aggregate the data.
We first combined task time and error metrics by normalizing them to $[0,1]$ and summing them together. The relative combined measure was computed using the performance of one condition minus the average performance of that participant. The resulting range is $[-2, 2]$, where a lower value indicates better performance. 

\subsubsection{Subjective Measures}
We administered a questionnaire based on prior research \cite{hoffman2019evaluating, rakita2020effects} to measure perceived \emph{predictability}, \emph{trust}, and \emph{fluency}.
Additionally, we employed NASA TLX \cite{hart1988development} to assess perceived workload. The six TLX subscale scores were averaged with equal weighting to calculate the overall TLX score.

 \subsection{Results}

Figure \ref{fig: results} and Table \ref{tab: statistical_results} summarize our results using one-way repeated-measures analyses of variance (ANOVA) to determine whether the designed hybrid frames had a significant effect. We used Tukey’s HSD test to make pairwise comparisons.
We found high inter-subject variability in our results, which matches prior studies that compare \controlframes{} \cite{hiatt2006coordinate, talha2019preliminary}. 
 
\subsubsection{Experiment 1: Pick-and-Place with VR Controller} Our results showed no significant difference but a medium-to-large effect size ($\eta^2 = 0.09$) in combined objective measures. We speculate that high inter-subject variability may be a cause for no significant difference.
As for self-reported measures, participants reported the robot with \emph{hybrid frame 1} to be significantly more fluent, trustworthy, and predictable, and to require significantly lower workload than the robot using \emph{robot frame}.
Comparing \emph{hybrid frame 1} with \emph{camera frame}, our results showed a significant difference only in the fluency metric and no significant difference in other self-reported metrics.
In the post-experiment interview, 16 out of 24 participants stated that they preferred the \emph{hybrid frame} condition. 3 participants preferred the \emph{camera frame} condition, and 2 participants couldn't separate the \emph{hybrid frame 1} and the \emph{camera frame} condition.
Being asked about the differences between conditions,  E1-P6 commented, \emph{``The [camera frame] seems very similar to [hybrid frame 1], just that when I move forward, [the camera frame] will move away from me and down.. If I want the robot to remain on an equal plane, I had to almost go like this'' (hand moves diagonally up)}. 
% \emph{"[The hybrid frame] is easier to remember which way the robot is going to move} (P1-11)"; 

As described in Section \ref{sec:case_studies:inperson}, to balance visual-motor misalignments and naturalness, \emph{hybrid frame 1} only has some rotational difference with the \emph{camera frame} along the camera pitch axis. Our study results indicated that most participants noticed the subtle distinctions between \emph{hybrid frame 1} and the \emph{camera frame}. Furthermore, the subtle distinctions created meaningful differences in user experience, although these differences did not produce statistically significant results in our experiment.

\subsubsection{Experiment 2: Pick-and-Place with Mouse} Our data showed that the interface with \emph{hybrid frame 2} significantly outperformed the interfaces using the \emph{robot frame} or \emph{camera frame} in the combined objective measure. When controlling the robot with \emph{hybrid frame 2}, participants reported significantly greater perceived fluency, trust, and predictability than the interface with the \emph{camera frame}. Although it took participants significantly more time to finish the task when using the \emph{robot frame}, there were no significant differences in subjective measures between the \emph{hybrid frame 2} and the \emph{robot frame} condition. After the study, 12 participants reported that they preferred the condition with \emph{hybrid frame 2}. E2-P5 commented: \emph{``The robot [with hybrid frame] moved kind of statically. Where in a short time I could master the movements. It seemed the easiest to use quickly.''} Meanwhile, 7 participants preferred the condition with the \emph{robot frame}, even though 6 of them were faster in the \emph{hybrid frame} condition. 

% P2-1 \emph{The [robot frame] robot feels like the controls are inverted and don't feel natural. }
%P2-3 \emph{The [robot frame] responded more from its own point of view. I had to think differently in order to operate that one since I was across from it. }
% \emph{P2-13 The [robot frame] robot had inverted controls which made it a bit tricky overall.}
%\emph{P2-16 The [robot frame] bot seemed to have the controls inverted}

\subsubsection{Experiment 3: Tracing with Mouse} Our results revealed that, across combined objective measure, predictability, trust, fluency, and TLX overall score, there were significant differences with the \emph{hybrid frame 3} being superior to the \emph{task frame}. Subjectively, 19 participants chose \emph{hybrid frame 3} as their preferred condition, while 5 participants preferred the \emph{task frame} condition.

\subsubsection{Summary} All three experiments showed the same tendency that the hybrid frames identified in our design space led to better performance and user perception compared to some commonly-used reference of frames. However, some of the differences were not statistically significant due to high inter-subject variability. We designed these hybrid frames based on the reasoning afforded by the design space and our experiment results match our predicted outcomes.
Our experiments confirm the predictions made by our design space, illustrating its utility as a conceptual tool in generating and evaluating \controlframes{}.

\section{Discussion}  \label{sec: discussion}

In this paper, we presented a design space for designers to understand the range of options in \controlframes{} and reason about the appropriateness of their designs for different scenarios. 
To showcase the utility of our design space, we organized existing \controlframes{} in our design space and conducted three case studies in which we designed \controlframes{} for various telemanipulation scenarios. These designed frames were evaluated in human subject experiments.
Our user evaluation showed that the choices and reasoning afforded by our design space allowed us to design and predict the performance and preference of \controlframes{} for various scenarios. 
Moreover, the subtle distinctions made visible by our design space created meaningful differences in user experience. The results highlighted the usability and generalizability of our design space.
Below, we discuss additional findings, implications for teleoperation systems, and the limitations of this work.

\emph{Individual Differences} --- Our experiment revealed considerable individual differences that may impact design choices. For example, our interview data showed that both mental rotation and perspective-taking strategies were used to address visual-motor misalignments, which matches the findings by Menchaca-Brandan \textit{et al.} \cite{menchaca2007influence}. % Using a mental rotation strategy, E1-P15 commented, \emph{``In the [robot] condition, I need to make some transformation.''} At the same time, E1-P23 used a perspective-taking strategy and commented, \emph{``[The robot frame] is like from the point of view of that screen (Sawyer robot's face). It's like you were on the other side of the desk...[The robot frame] is the one I first experienced. I got used to that, so it just got stuck in my brain a little.''}  It seems hard for participants to adapt perspective-taking strategies to other conditions, \emph{``[At the start of the camera frame condition], I feel like I couldn't find my position''} (E1-P16).
For some participants who used perspective-taking strategies, it seemed hard for them to adapt to a new \controlframe{}. E1-P23 commented, \emph{``[The robot frame] is like from the point of view of that screen (Sawyer robot's face)... [The robot frame] is the one I first experienced. I got used to that, so it just got stuck in my brain a little.''}. Similarly, E1-P16 commented, \emph{``[At the start of the camera frame condition], I feel like I couldn't find my position.''} 
% Interestingly, participant E1-P28 analogized robot control mappings to touchpad scroll direction. Compared to \emph{natural scrolling} (content tracks finger movement), the participant is more habituated to \emph{reverse scrolling} (window tracks finger movement, so the content direction is reverse). Therefore, in our robot study, by moving the controller to a direction, the participant expected to see a reverse robot direction on screen. Consequently, the participant reported the \emph{robot frame} to be the most intuitive condition. This suggests that people's mapping habits may have been shaped by other technology products (e.g., Mac OS and Windows' default scrolling directions are opposite).
Moreover, one participant who preferred the \emph{robot frame} in experiment 1 stated the reason as being habituated to reverse scrolling settings when using a touchpad. 
These individual differences suggest that a telemanipulation system may benefit from personalized mappings (\textit{e.g.}, \cite{li2020learning, pierce2012data}).

% P1-15 "In the [hybrid mapping and camera frame] condition, I do not need to think which direction I want to go... but in the [robot] condition, I need to make some transformation."
% P2-20 commented \emph{"The [robot frame] interacted as if my input were first person from the robot POV."} 
% \emph{P1-12: I feel like the second [hybrid] and third [camera] condition didn't move as I expected. When I push left, it goes right; when I push forward, it goes back}. Note that the participant was describing robot movement with respect to the robot frame. 
% \emph{P2-5: The [robot frame] robot seemed to move naturally with my mouse movements... The [hybrid mapping] robot seemed to reverse my mouse movements.}
% \emph{P2-7: Honestly, once I got used to how they moved, I don't think there was much of a preference.}
%Most people can't tell the differences between the camera frame and hybrid frame. 
 %Some people feel hybrid control is too sensitive while others sense not sensitive enough.

\emph{Implications} ---
Besides mapping user inputs to some commonly used reference frames in the remote environment (\textit{e.g.}, \emph{robot frame}), there are many other possible \controlframes{} for telemanipulation. Designers can tailor \controlframes{} for their various scenarios by choosing meaningful directions from our design space.
To reason about the suitability of \controlframes{}, prior research \cite{dejong2004improving} emphasized the reduction of mental workload caused by visual-motor misalignments; however, our experiments showed that some visual-motor misalignments along the pitch axis (in the \emph{hybrid frames} in pick-and-place experiments) do not increase mental workload. These results match findings in the ergonomics community \cite{ellis2012human}.
Besides minimizing visual-motor alignments, designers should take into consideration other design criteria provided in this work, \textit{i.e.}, naturalness and task semantics. 

\emph{ Limitations \& Future Work} --- Our work has some limitations that must be addressed by future work. First, while our design space allows us to choose \controlframes{} for different scenarios, we cannot make claims about its comprehensiveness (as defined by Kerracher and Kennedy \cite{kerracher2017constructing}). 
We summarized the design criteria based on prior work and our design experience but our design space may be extended by additional criteria. 
Moreover, directions decomposed from some learning-based mappings \cite{li2020learning, losey2020controlling, pierce2012data} may provide insights to extend our design space.
Meanwhile, our design criteria may be used to reason about the suitability of learning-based mappings or encoded in learning-based methods to generate more intuitive mappings.
Second, while \controlframes{} map both translational inputs and rotational inputs to the robot, we only assess \controlframes{} on translational inputs in our human subject experiments. Future work should evaluate \controlframes{} on manipulation tasks that involve both translational and rotational movements. 
Besides selecting a \controlframe{}, future work should explore alternative mapping approaches, \textit{e.g.}, a mapping from translational inputs to rotational robot movements \cite{campeau2018intuitive}. 
Third, while our approach applies to mode-switched interfaces, future work should extend our framework to consider the costs and benefits of changing \controlframes{}.
Fourth, in this work, all \controlframes{} were evaluated on low-latency teleoperation systems. Future work should explore \controlframes{} for high-latency teleoperation systems where operators get delayed feedback, introducing additional challenges for the design of \controlframes{}.
Fifth, while our paper provides a design space to assist telemanipulation system designers, there are several trade-offs designers need to consider, such as choosing a task-specific or a general design (described in \cref{sec:criteria}), prioritizing either user experience or task performance, or tailoring \controlframes{} for expert or non-expert users. 
Last, in the post-experiment interview in experiment 1, some participants complained about the depth ambiguity caused by the single, static camera setting. Future work should investigate \controlframes{} under multiple or dynamic camera views (\textit{e.g.}, \cite{praveena2022understanding}).

\emph{Conclusion} --- In this paper, we present our design space that articulates the range of options in \controlframes{} and provides design criteria for designers to reason about the suitability of their designs for various telemanipulation scenarios. We believe that the contribution of our design space with respect to designing and predicting suitable \controlframes{} will enhance and facilitate the input mapping design process for future telemanipulation systems. 

\bibliography{root}
\bibliographystyle{IEEEtran}

\end{document}